# A Method of Detecting End-To-End Curves of Limited Curvature

E.I. Panfilova[2,7], M.A. Aliev[1,2], I.A. Kunina[2,3], V.V. Postnikov[3,4,5], D.P. Nikolaev[3,6]

[1] Federal Research Center "Computer Science and Control" of Russian Academy of Sciences, Moscow, Russia
[2] Smart Engines Service LLC, Moscow, Russia
[3] The Institute for Information Transmission Problems of Russian Academy of Sciences, Moscow, Russia
[4] Cognitive Technologies, Moscow, Russia
[5] National University of Science and Technology "MISIS", Moscow, Russia
[6] Moscow Institute of Physics and Technology, Dolgoprodny, Russia
[7] V. A. Trapeznikov Institute of Control Sciences of Russian Academy of Sciences, Moscow, Russia

## ABSTRACT

In this paper we consider a method for detecting end-to-end curves of limited curvature like the k-link polylines with bending angle between adjacent segments in a given range. The approximation accuracy is achieved by maximization of the quality function in the image matrix. The method is based on a dynamic programming scheme constructed over Fast Hough Transform calculation results for image bands. The proposed method asymptotic complexity is $O(h \cdot (w+h/k) \cdot log(h/k))$, where $h$ and $w$ are the image size, and $k$ is the approximating polyline links number, which is an analogue of the complexity of the fast Fourier transform or the fast Hough transform. We also show the results of the proposed method on synthetic and real data.

## 1. INTRODUCTION

The detection of simple figures defined by a finite number of parameters is familiar subproblem in different image processing and analysis problems. Most methods of detecting such figures based on detectors of boundary points, which are modeled as edge or ridge. Unfortunately, objects boundaries are noisy in the image because of optical systems imperfections. That's why the result of boundary detection algorithm not always can be matched to real object. One of the methods to detect simple figures in noisy images is Hough Transform (HT).

HT is a well-known and widely used tool in image processing and analysis. The classical HT was proposed to detect straight lines in an image [1]. The main idea of HT is to estimate the number of points lying on every line in the image. Afterward HT was generalized to detect different shapes in the image - ellipses, circles, etc. [2, 3]. However, the computational complexity increases with the increasing number of target shape parameters.

At present there are a lot of different modifications of HT. Their overview can be found in [4]. For example, Randomized Hough transform [5] and Probabilistic Hough transform [6] are less computationally expensive, Hough transform 3D [7] detects angles, segments and polylines. However, the main purpose of the HT and its modifications remains lines detection [8, 9], due to the fact that the line is defined by two parameters, and therefore the calculation of the HT for this form is asymptotically faster.

The fast Hough transform (FHT) [10] is also one of Hough transform modifications. It is as well noise-resistant, and its computational complexity is $O(h \cdot w \cdot log(h))$, where $h$ and $w$ are source image height and width. Moreover, it was proved in [11] that such computational complexity is asymptotically optimal. FHT is applied not only for lines detection, but also for the other problems. For example, in [12] FHT is used for calculation of M-estimation in the orthogonal linear regression task. Also, in [13] a new fast exact algorithm for optimal linear separation of 2d distribution is proposed. In [14] FHT is used to estimate the overall curvature of the lines without actually detecting them in the image. The algorithm we proposed uses FHT to detect end-to-end curves of limited curvature in the form of polylines. Unlike HT3D, it detects polylines with certain features, like bounded bending angle, and it has less computational complexity.

The problem of detection end-to-end curves limited curvature is in high demand. An example is lane markings detection for map-relative unmanned ground vehicle localization [15, 16, 17] or building a lane map [18]. Similar problem is solved by unmanned aerial vehicles, but they detect roads [19] in optical aerial images and it is also take place in system for controlling trajectories of the agricultural combine harvester [20]. Another example is finding track of each wheel in video sequence for use in vision-based automatic vehicle classifiers with narrow field of vision [21], or estimation tracks of other moving objects [22]. One more example is teeth shape detection in an orthopantomogram for people identification [23, 24]. Curves detection can be also useful for a document localization in the image. The papers [25, 26] consider cases when the document image has the shape of a quadrangle or is very close to it, that is, the document boundaries in the image are straight lines. If the document has been bent, the document boundaries are curves and the proposed algorithms will fail.

Curves detection is not a new problem. It is often solved using edge detection algorithms such as Canny edge detector (original implementation [27] or maybe in the form of neural networks [28]) and others. Such an approach can be seen in the papers [29, 30]. However, if the target line could be modeled as a line passing through an area with constant brightness, it leads to doubled boundaries. To detect such lines ridge detectors can be used [31, 26]. However, all these methods detect curves is without curvature restrictions or, in other words, can detect lightning-like curves, which are not always satisfy the problem conditions. The algorithm in paper [32, 19] is based on the window FHT. The found segments are approximated into straight or poly- lines bounded by bending angle. This approach however has high computational complexity.

In this paper we consider a method for detecting end-to-end curves of limited curvature. The method is based on a dynamic programming scheme constructed over Fast Hough Transform calculation results for image bands.

The paper is structured as follows: in Section 2 we formulate the main problem, and then it is solved by sequentially solving subproblems in Sections 2.1 - 2.3. In Section 3 we show the proposed algorithm results on synthetic and real data. Finally, conclusions are given in Section 4.

## 2. PROPOSED POLYLINES DETECTION ALGORITHM

Let us introduce several definitions.

1. The segment **inclination angle** (see Fig. 1a) is the angle between the axis $Oy$ and the segment, measured clockwise. If the angle is greater than 90°, then it is considered counterclockwise from the Oy axis and is assigned a negative sign.
2. A **segment** or a straight **line** is called **mostly vertical** (**MVS** or **MVL**) if its inclination angle is in the range $[-45°; 45°]$[10].
3. **Image horizontal $k$-division** is division along the $Oy$ axis into $k$ equal height bands.
4. **$k$-link mostly vertical uniform polyline (MVUP)** is a polyline in the image with a horizontal $k$-division, all the links of which are MVS and the ends of the links lie on the horizontal boundaries of the image bands.
5. The polyline **bending angle** at a polyline vertex (see Figure 1b): Let there be two adjacent links $AB$ and $BC$, then the bending angle is the angle between the vectors $\overrightarrow{AB}$ and $\overrightarrow{BC}$.

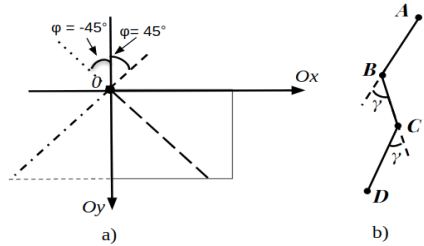

Figure 1. The definitions of the segment inclination angle $\varphi$ and the polyline bending angle $\gamma$. $AB$, $BC$, $CD$ - polyline segments, thus $\angle(\overrightarrow{AB}, \overrightarrow{BC})$ and $\angle(\overrightarrow{BC}, \overrightarrow{CD})$ are bending angles by definitions.

**Main problem statement. Detection of curve approximating polyline with limited bending angle.**

Suppose we have an image of width $w$, height $h$, with a horizontal $k$-division (see Fig. 2). It is known that the image has a curve $l$, along which the values are large, and

1. the inclination angle $\alpha$ of the tangent to $l$ is in the range $[-45°, 45°]$;
2. $\omega_\alpha\left(\frac{h}{k}\right) \leq \gamma_{max}$, where $\omega_\alpha$ – stands for modulus of continuity $\alpha(y)$;
3. $l$ runs from top to bottom of the image.

There is also uncorrelated spiky noise in the image.

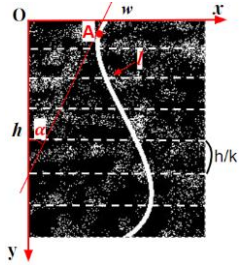

Figure 2. An example of an input image. A tangent to $l$ is drawn through the point $A$.

The goal is to find the MVUP, which approximately minimizes the mean-square deviation from $l$ by the $Ox$ axis.

The real example where detecting end to end curves of limited curvature can be helpful is at figure 3. We see a bent document on the complex background (see Fig. 3a). The boundaries of background objects have bigger pixel values than document boundaries in the morphological gradient of reduced channel document photo (see Fig. 3b), that's why it is necessary to detect bent document boundaries like end-to-end extreme polylines with small bending angle. In figure 3, we also can see the detecting result of MVUP (see Fig. 3c) versus MVUP with limited bending angle (see Fig. 3d).

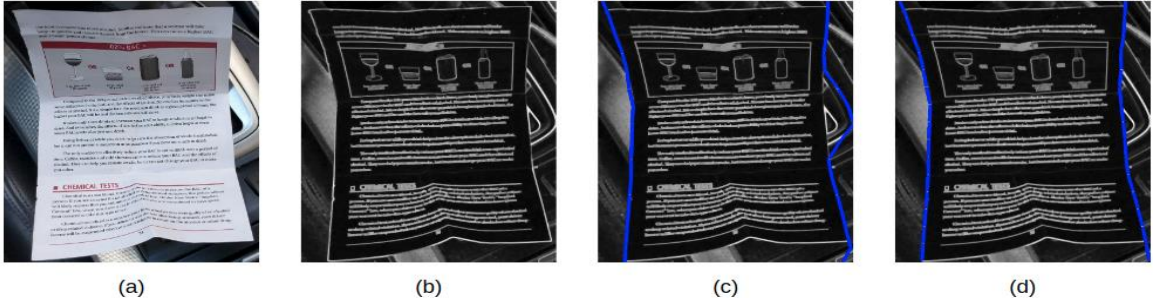

Figure 3. (a) A photo of bent document was taken from [33] open dataset. (b) Morphological gradient of reduced channel document photo. (c) Result of 2 extreme MVUP detection. (c) Result of detection 2 extreme MVUP limited curvature, the maximum of bending angle is 10°;

Now we consider a few subproblems to solve the main problem.

## 2.1 Subproblem 1. Calculation the pixel value sum along the MVUP.

Suppose we have an image of width $w$, height $h$, with a horizontal $k$-division. We need to create the quick response algorithm to a series of queries like "Calculate the pixel value sum along a given MVUP".

**Algorithm**: Let's extend each band to the right by $\frac{h}{k}$, filling it with zero valued pixels, as suggested in [34]. For each of $k$ extended bands we compute the Hough-image for mostly vertical lines using the FHT [10]. We denote them as $H_0, H_1, \ldots, H_{k-1}$. By definition the Hough-image for mostly vertical lines contains pixel value sums along all mostly vertical segments, whose endpoints lie on the image sides. Suppose we need to calculate the pixel value sum along the MVUP with coordinates $\left[(x_0; 0), \left(x_1, \frac{h}{k}\right), \left(x_2, \frac{2h}{k}\right), \ldots, (x_k, h)\right]$. Each polyline segment $\left[\left(x_i, \frac{i \cdot h}{k}\right); \left(x_{i+1}, \frac{(i+1) \cdot h}{k}\right)\right]$, $i \in [0, k-1]$ corresponds to a point in the Hough-image $H_i$ with coordinates (see [34]) $(x_i, shift)$, where

$$shift = \frac{h}{k} - x_{i+1} + x_i \qquad (1)$$

the value of which is equal to the pixel value sum along this segment.

Thus $\sum_{i=0}^{k-1} H_i\left(x_i, \frac{h}{k} - x_{i+1} + x_i\right)$ is the required sum along the polyline $\left[(x_0; 0), \left(x_1, \frac{h}{k}\right), \left(x_2, \frac{2h}{k}\right), \ldots, (x_k, h)\right]$.

**Asymptotic complexity.**

The complexity of the Hough-image calculation for mostly vertical lines of the band is $O\left(\left(w + \frac{h}{k}\right) \cdot \frac{h}{k} \cdot \log\left(\frac{h}{k}\right)\right)$ [34]. Since there are $k$ bands, the precalculation for the entire image requires $O\left(\left(w + \frac{h}{k}\right) \cdot h \cdot \log\left(\frac{h}{k}\right)\right)$ operations.

## 2.2 Subproblem 2. Detection of extreme polyline through a given point.

Let we have an image of width $w$, of height $h$, with a horizontal $k$-division, and a point $P$ on one of the bands horizontal boundaries. We need to create an algorithm which detects the extreme MVUP (in terms of the maximum pixel value sum along it) passing through a given point $P$.

**Algorithm**: Let the given point be on the upper border of the band $j$, $j \in [0; k-1]$. If the point were on the lower band border, it would mean that it is on the upper border of the $j+1$ band, so this assumption does not restrict generality. Note that the band upper border will have the smallest $y-$ coordinate in the image coordinate system. Let $\left(x_j, j \cdot \frac{h}{k}\right)$ be the coordinates of point $P$. Now let's detect an extreme polyline.

Let's begin with detection of the extreme polyline part which starts at the point $P$ and passing through the bands $j, \ldots, k-1$. To do this, we compute the Hough-image for mostly vertical lines for bands $j, \ldots, k-1$. Denote them as $H_j, \ldots, H_{k-1}$ (see Fig. 4).

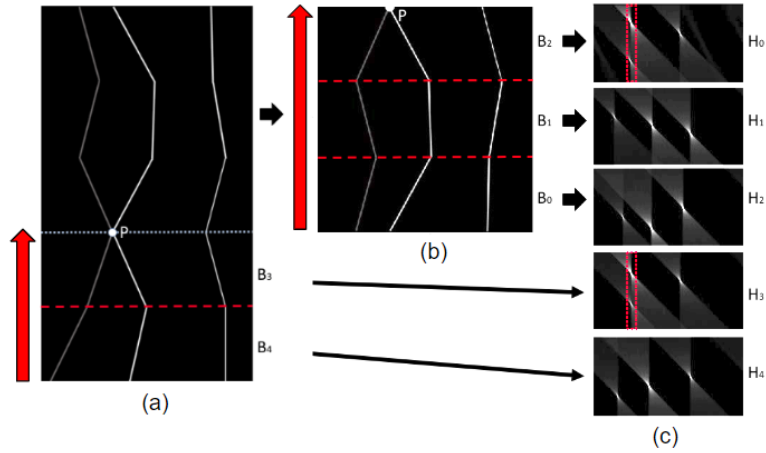

Figure 4. (a) Input image with 5-division, point $P$ with coordinates $\left(x_3, \frac{3h}{k}\right)$, red arrows indicate the polyline calculation direction; (b) Flipped image top part; (c) Input image bands Hough-images; red columns correspond to segments coming out from the point $P$.

For better understanding the following part of the algorithm, note a few points:

1. By the point coordinates $(x^*, shift)$ in the Hough-image for mostly vertical lines it is possible to compute coordinates of the corresponding segment ends in the source image. One of the end coordinates is $(x^*, 0)$, and the second one is $(x^{**}, h)$, where $h$ is the source image height and
$$x^{**} = h - shift + x^* \qquad (2)$$
This formula follows from formula (1).
2. The $x^*$- column in the Hough-image for mostly vertical lines corresponds to the MVS coming up from one point $(x^*, 0)$ (see fig. 4c).

Now we step by step detect the polyline bottom-up to the point $P$ (see Fig. 4, the red arrows indicate the calculation direction) in the following way (Notice that we use dynamic programming principles in this algorithm.):

From $i = k - 1$ to $j + 1$:
1. find and remember the maximum coordinates in the Hough-image columns

$$ArgMax_i = \left\{(x_{i,t}, shift_{i,t}): H_i(x_{i,t}, shift_{i,t}) = max_{y=0}^{\frac{2h}{k}} H_i(t, y), t \in \left[0, w + \frac{h}{k} - 1\right]\right\}$$

2. For each $(x_0, sh) \in H_{i-1}$:

$$M = H_i\left(x_{i,\frac{h}{k}+x_0-sh}, shift_{i,\frac{h}{k}+x_0-sh}\right) \quad \text{— pixel value sum along the extreme segment} \quad (3)$$

coming up from the point $\left(\frac{h}{k} + x_0 - sh, i \cdot \frac{h}{k}\right)$

$$H_{i-1}(x_0, sh) \mathrel{+}= M \tag{4}$$

As a result, after step $i$ is completed, the Hough-image $H_{i-1}$ will store the pixel value sums along the extreme polylines (with the maximum pixel value sum along polyline of all possible polylines), starting in the band $i - 1$ and ending in the band $k - 1$. In the $B_j$ band we will analyze only those polylines that come up from a given point $\left(x_j, j \cdot \frac{h}{k}\right)$.

For this purpose, we find maximum coordinates in the columns of the Hough-image $H_{j+1}$, and then update the values of the $x_j$-column in Hough-image $H_j$ by the formula **(4)**. As a result, the maximum in the $x_j$-column in the Hough-image $H_j$ will correspond to the required extreme polyline part. Let $(x_j, shift_{j,x_j})$ be the maximum coordinates, then the coordinates of required extreme polyline part is

$$\left\{\left(x_{j+n}, shift_{j+n,x_{j+n}}\right) : n \in [0, k-j]\right\}, \tag{5}$$

where $x_{j+n} = \frac{h}{k} + x_{j+n-1} - shift_{j+n-1,x_{j+n-1}}$ — by formula (2) and $shift_{k,x_{k-1}} = h$

The upper part of the extreme polyline can be found by flipping over the image part corresponding to bands $0, \ldots, j - 1$ about the Ox axis (see Fig. 3b) and then use the described above algorithm.

**Asymptotic complexity.**

As in the previous subproblem, the precalculation will be done in $O\left(\left(w + \frac{h}{k}\right) \cdot h \cdot \log\left(\frac{h}{k}\right)\right)$ operations, and the extreme polyline detection in $O\left(k \cdot \left(\left(w + \frac{h}{k}\right) \cdot \frac{2h}{k} + \frac{2h}{k} \cdot \left(w + \frac{h}{k}\right)\right)\right) = O\left(h \cdot \left(w + \frac{h}{k}\right)\right)$, because at each step both substeps require $O\left(\left(w + \frac{h}{k}\right) \cdot \frac{2h}{k}\right)$ operations.

Let's complicate the problem further.

### 2.3 Subproblem 3. Detection of extreme polyline through a given point with limited bending angle.

Let we have an image of width $w$, of height $h$, with a horizontal $k$-division, and a point $P$ on the horizontal boundary of one of the bands. We need to create an algorithm which detects the extreme MVUP (in terms of the maximum pixel value sum along it) passing through a given point $P$ and with bending angle restriction — $[0, \gamma_{max}]$, where $\gamma_{max} \in [0; 90°]$.

**Algorithm**:

The algorithm is similar to the previous one. But at each step we will look for an extreme polyline continuation for the processing segment, which will satisfy the restriction of the bending angle. This means that in the Hough-image we will not consider the entire column, but only its sub-range.

For better algorithm understanding, let's notice some points. Let we have the image of width $w$, of height $2h$, with a horizontal 2-division. Hough-image for MVL is calculated for each band: $H_1$ — for the top band and $H_2$ — for the bottom one. Then:

1. Let $\varphi$ be the inclination angle of the segment, then we can calculate its Hough-image coordinate along the ordinate axis $shift = h + h \cdot tg(\varphi)$.
2. Suppose we have two segments in the upper and lower image bands coming up from the same point on the bands' border (see fig. 5). By $(x_1, sh_1)$ denote upper segment Hough-image, then $(h - sh_1 + x_1; sh_2)$ is the the Hough-image of the lower segment, $\alpha = arctg\left(\frac{sh_1 - h}{h}\right)$ is the upper segment inclination angle and

$\beta = arctg\left(\frac{sh_2-h}{h}\right)$ is the lower segment inclination angle. Then the smaller angle between segments $\gamma_{min}$ is $min\{\pi + \beta - \alpha; 2\pi - (\pi + \beta - \alpha)\}$. Hence the 2-links polyline bending angle $\gamma = |\alpha - \beta|$.

3. Suppose $(x_1, sh_1)$ is some point of the Hough-image $H_1$. Let's find all points in $H_2$, whose bending angle formed by their preimages with the preimage $(x_1, sh_1)$ will satisfy the acceptable range $[0, \gamma_{max}]$. The target points coordinates by the abscissa axis are $h - sh_1 + x_1$ (see formula (2)). To find the coordinate by the ordinate axis, let's solve 2 system on the bending angle restriction:

$$\begin{cases} 0 \leq \alpha - \beta \leq \gamma_{max} \\ \beta \in \left[-\frac{\pi}{4}; \frac{\pi}{4}\right] \end{cases} \quad or \quad \begin{cases} 0 \leq \beta - \alpha \leq \gamma_{max} \\ \beta \in \left[-\frac{\pi}{4}; \frac{\pi}{4}\right] \end{cases}$$

where $\alpha$ and $\beta$ accordingly stand for the inclination angles of pre-images of the given and target Hough-image points. Hence, we can get the restrictions of the coordinate by the ordinate axis for the points we search:

$$h + h \cdot tg\left(max\left\{-\frac{\pi}{4}; \alpha - \gamma_{max}\right\}\right) \leq shift \leq h + h \cdot tg(\alpha) \quad (6)$$

or

$$h + h \cdot tg(\alpha) \leq shift \leq h + h \cdot tg\left(min\left\{\frac{\pi}{4}; \alpha + \gamma_{max}\right\}\right) \quad (7)$$

where $\alpha = arctg\left(\frac{sh_1-h}{h}\right)$

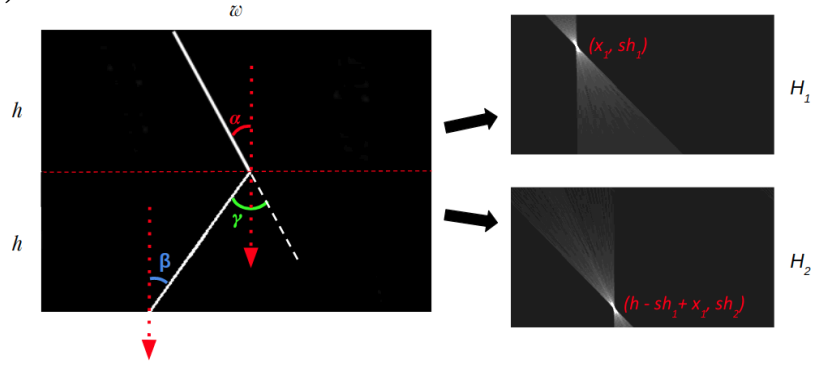

Figure 5. Image with horizontal 2-division, of width $w$ and height $2h$. $H_1, H_2$ stand for bands Hough-images. $\alpha, \beta$ — inclination angles, $\gamma$ — bending angle

Let's return to the problem solution. It is the same as the solution of subproblem 2, with the only difference that at first substep of the step-by-step process we will calculate a data structure for each column that can quickly respond to the query "Find the subrange maximum". Let's call it $RangeMaxQ_i^{x_0}$ for step $i$ and $x_0$- column. The second step is this:

For each $(x_0, sh) \in H_{i-1}$ :

1. From inequalities **(6)** and **(7)** we find Hough-images of segment satisfied the bending restriction. These are the sub-range $[sh^1; sh^2]$ and $[sh^3; sh^4]$ of the $H_i$ column with the coordinate — $\frac{h}{k} - sh + x_0$.
2. $M = max(RangeMaxQ_i([sh^1; sh^2]), RangeMaxQ_i([sh^3; sh^4]))$, the index of this maximum is denoted as $shift_{i,\frac{h}{k}+x_0-sh}$. Hence $M$ is the pixel value sum along the extreme segment coming up from the point $\left(\frac{h}{k} + x_0 - shift, i \cdot \frac{h}{k}\right)$ satisfying the bending restriction.
   $H_{i-1}(x_0, sh) += M$

Thus, the polyline coordinates will be the same as in the subproblem 2. Precisely, the formula of the lower polyline part from point $P$ is the formula **(5)**.

**Asymptotic complexity.**

As in the previous subproblem, the precalculation will be done in $O\left(\left(w + \frac{h}{k}\right) \cdot h \cdot log\left(\frac{h}{k}\right)\right)$ operations, and the asymptotic complexity of extreme polyline detection depends on what data structure we use to answer the maximum element range query. If we use Sparse table [35], the precalculation for each Hough-image column will take

$O\left(\frac{h}{k} \cdot log\left(\frac{h}{k}\right)\right)$, and there are $\left(w + \frac{h}{k}\right)$ columns in total. Respond to the maximum element query is $O(1)$. Thus, the extreme polyline detection will be done in $O\left(\left(\left(w + \frac{h}{k}\right) \cdot \frac{h}{k} \cdot log\left(\frac{h}{k}\right) + \left(w + \frac{h}{k}\right) \cdot \frac{h}{k}\right) \cdot k\right) = O\left(\left(w + \frac{h}{k}\right) \cdot h \cdot log\left(\frac{h}{k}\right)\right)$. If we use Segment Tree [36], the precalculation for each Hough-image column will take $O(N), N = 2^{\widehat{log_2 \frac{h}{k}}}$, where $\hat{x}$ stands for rounding up. A response to request will take $O(log(N))$. Therefore, the extreme polyline detection will take

$O\left(\left(\left(w + \frac{h}{k}\right) \cdot \frac{h}{k} + \left(w + \frac{h}{k}\right) \cdot \frac{h}{k} \cdot log\left(\frac{h}{k}\right)\right) \cdot k\right) = O\left(\left(w + \frac{h}{k}\right) \cdot h \cdot log\left(\frac{h}{k}\right)\right)$ operations.

As we can see, the whole asymptotic complexity is equal for these two data structures. The difference is that precalculation for sparse table is more computationally complex than for segment tree. On the contrary, the answer for range maximum query is quicker.

Now let's return to the main problem - detection of curve approximating polyline with limited bending angle.

**Solution to the main problem**. We solve the problem by calculating the Welsh M-estimation [37] of a $k$-link MVUP with the bending angle restriction. In fact, the Welsh error function is close to the mean-square deviation near 0, but in contrast it is robust to uncorrelated spiky noise.

**Algorithm**:
1. Convolute the input image line by line with one-dimensional Gaussian, denote the resulting image $I_{gaussian}$.
2. Using the solution of subproblem 3, let's find the extreme MVUPs in the $I_{gaussian}$ passing through the points with coordinates $(0,0), \ldots, (w-1, 0)$, and whose bending angle $\gamma$ will satisfy the restriction: $\gamma \in [0; \gamma_{max}]$ since $\omega_\alpha = sup(|\gamma|) \leq \gamma_{max}$.
3. Choose from the found polylines the one, the sum of pixel values along which will be maximal. Thus, we find the target polyline.

The extreme polyline in the image smoothed by Gaussian will correspond to the Welsh M-estimation for the MVUP approximating the curve in the original image. For the first time this approach was used in [38].

**Asymptotic complexity.**

As in subproblem 3, the asymptotic complexity of polyline detection is $O\left(\left(w + \frac{h}{k}\right) \cdot h \cdot log\left(\frac{h}{k}\right)\right)$, and the precalculation complexity is $O\left(h \cdot w \cdot w_{gaussian} + \left(w + \frac{h}{k}\right) \cdot h \cdot log\left(\frac{h}{k}\right)\right) = O\left(\left(w + \frac{h}{k}\right) \cdot h \cdot log\left(\frac{h}{k}\right)\right)$, where $w_{gaussian}$ is the convolution window size.

### 3. APPLICATION OF THE ALGORITHM TO SINGLE IMAGES

The proposed algorithm was tested on single images of synthetic and real data and showed its practical applicability. The results can be seen at the figure 6 and 7. Red arrows show the direction of the calculation for MVUP.

Figure 6 examples (a)-(c) represent the algorithm results on synthetic data, and figure 6 examples (d), (e) and figure 6 - on real data. In fig. 6 examples (a)-(c) display one best polyline. In fig. 6 (d), fig. 7 (a) two extreme, different from each other, polylines are displayed. The original images can be obtained from the link [39]. Image (d) is taken from an open dataset [33] from the article [40].

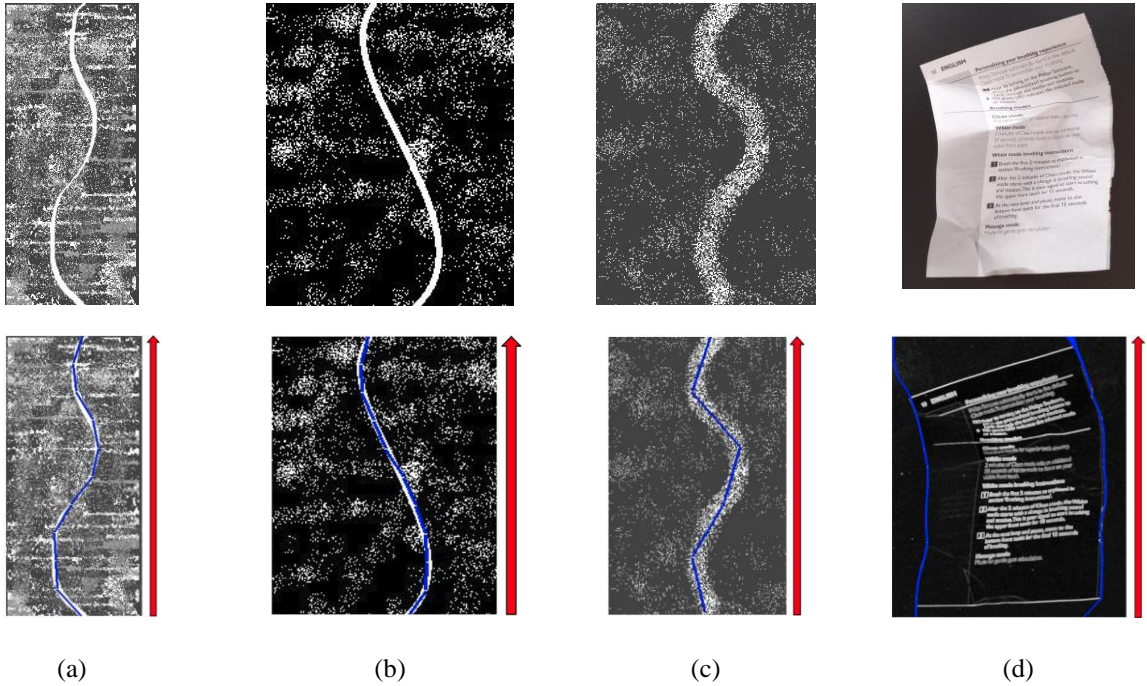

Figure 6. Examples of the results of the proposed algorithm. Top row - input images, bottom row - algorithm result.

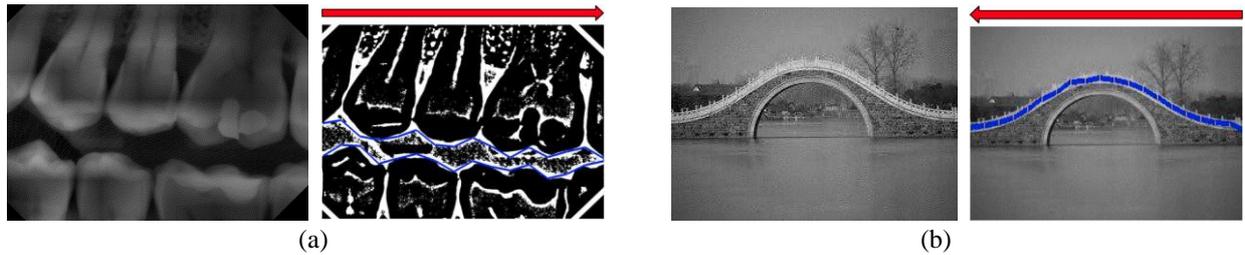

Figure 7. Mostly horizontal polyline detection. Examples of the algorithm results. Top row - input images, bottom row - algorithm result.

## 4. CONCLUSION

In this paper we propose an algorithm for detection the end-to-end curves of limited curvature, extreme in the sense of the pixels value sum. The algorithm result is a set of MVUP with a restriction on the bending angle, which are approximations of the required curves.

The presented algorithm uses FHT and dynamic programming method. Such an approach reduces the overall computational complexity of the original problem. Asymptotic complexity of the algorithm precalculation and the target polyline detection is $O\left(\left(w + \frac{h}{k}\right) \cdot h \cdot log\left(\frac{h}{k}\right)\right)$.

To solve the problem there were considered sequentially 3 more and more complicated subtasks: the answer to the query "Calculate the sum of pixels value along the given MVUL"; detection of extreme MVUP (i.e. with the maximum pixels value sum) passing through a given point in the image; and detection of the extreme MVUP with a bending angle restriction passing through a given point in the image. The proposed algorithm was tested on single synthetic and real images and showed its practical applicability. In further work the authors plan to obtain a quantitative estimation of the algorithm and improve it to find more complex structures – such as forks.

## ACKNOWLEDGMENTS

This work is partially financially supported by Russian Foundation for Basic Research (projects 19-29-09075, 18-29-26017)